\documentclass[letterpaper, 10 pt, conference]{ieeeconf}  % Comment this line out if you need a4paper
\IEEEoverridecommandlockouts                              % This command is only needed if
\usepackage{graphics}    % for pdf, bitmapped graphics files
\usepackage{times}       % assumes new font selection scheme installed
\usepackage{amsmath}     % assumes amsmath package installed
\usepackage{amssymb}     % assumes amsmath package installed
\usepackage{graphicx}
\usepackage{algorithm}
\usepackage[noend]{algpseudocode}
\usepackage{hyperref}
\usepackage{listings}
\usepackage{xcolor}
\usepackage{multirow}
\usepackage{placeins}

\def\figref#1{Fig.~\ref{#1}}
\def\tabref#1{Tab.~\ref{#1}}
\def\eqref#1{Eq.~(\ref{#1})}

\def\listref#1{List.~\ref{#1}}

\newcommand\class[1]{\texttt{#1}}

\usepackage{fancyhdr}
\title{\LARGE \bf Bonnet: An Open-Source Training and Deployment Framework for Semantic Segmentation in Robotics using CNNs}

\author{Andres Milioto \and Cyrill Stachniss% <-this % stops a space
  \thanks{All authors are with the University of Bonn, Germany. This work has 
  partly been supported by the EC under the grant number H2020-ICT-644227-Flourish. 
  }%
}

\begin{document}
\maketitle
\thispagestyle{fancy}
\pagestyle{fancy}

%%%%%%%%%%%%%%%%%%%%%%%%%%%%%%%%%%%%%%%%%%%%%%%%%%%%%%%%%%%%%%%%%%%%%%%%%%%%%%%%
\begin{abstract}
  %
  % WHY is it relevant?
  %
  The ability to interpret a scene is an important capability for a robot that is supposed to interact
  with its environment. The knowledge of \emph{what} is in front of the robot is, for example,
  relevant for navigation, manipulation, or planning.
  Semantic segmentation labels each pixel of an image with a class label and thus provides
  a detailed semantic annotation of the surroundings to the robot. Convolutional neural 
  networks (CNNs) are popular methods for addressing this type of problem.
  The available software for training and the integration of CNNs for real robots, however, is 
  quite fragmented and often difficult to use for non-experts, despite the availability 
  of several high-quality open-source frameworks for neural network implementation and training.
   %
  % WHICH PROBLEM 
  %
  In this paper, we propose a tool called Bonnet, which addresses this
  fragmentation problem by building a higher abstraction that is specific for
  the semantic segmentation task. It  provides a modular approach to simplify the 
  training of a semantic segmentation CNN independently of the used 
  dataset and the intended task. Furthermore, we also address the deployment on a real robotic platform.
  Thus, we do not propose a new CNN approach in this paper. Instead, we provide a
  stable and easy-to-use tool to make this technology more approachable in the context of
  autonomous systems. In this sense, we aim at closing a gap between computer vision research and its use in robotics research. 
  %
  % HOW 
  %
  We provide an open-source codebase for training and deployment.
  The training interface is implemented in Python using TensorFlow and the deployment
  interface provides a C++ library that can be easily integrated in an existing robotics
  codebase, a ROS node, and two standalone applications for label prediction in images and videos.
\end{abstract}

%%%%%%%%%%%%%%%%%%%%%%%%%%%%%%%%%%%%%%%%%%%%%%%%%%%%%%%%%%%%%%%%%%%%%%%%%%%%%%%%
\section{Introduction}
\label{sec:intro}

%% WHY 

Perception is an essential building block of most robots. Autonomous
systems need the capability to analyze their surroundings in order to safely and
efficiently interact with the world.  Augmenting the robot's camera data with
the semantic categories of the objects present in the scene, has the potential
to aid
localization~\cite{armagan2017jurse,atanasov2016ijrr,poschmann2017ECMR},
mapping~\cite{Khanna2015etfa,sunderhauf2015arxiv}, path planning and
navigation~\cite{drouilly2015ICRA,zhao2015icac},
manipulation~\cite{Blodow2011iros,Schwarz2017IJRR}, precision farming~\cite{lottes2016icra,milioto2018icra,milioto2017isprsannals}
as well as many other tasks and
robotic applications. Semantic segmentation provides a pixel-accurate category
mask for a camera image or an image stream.  The fact that each pixel in the
images is mapped to a semantic class,   allows the robot to obtain  a
detailed semantic view of the world around it and aids to the understanding
the scene.

%% WHICH PROBLEM 

Most methods, which represent the current state of the art in semantic
segmentation, use fully convolutional neural networks. The success of
neural networks for many tasks from machine vision to natural language
processing has triggered the availability of many high-quality
open-source development and training frameworks such as TensorFlow~\cite{abadi2016arxiv}, Caffe~\cite{jia2014arxiv}, or Pytorch~\cite{paszke2017automatic}. 
Even though these frameworks have simplified the development of new networks and
the exploitation of GPUs dramatically, it is still non-trivial for a novice to
build a usable pipeline from training to deployment in a robotic platform.
Companies such as NVIDIA and Intel have furthermore developed custom accelerators 
such as TensorRT or the Neural Compute SDK. Both use graphs created with
TensorFlow or Caffe as inputs and transform them into a format in which inference can be
accelerated by custom inference hardware.  As with the other frameworks, their learning
curve can be steep for a developer that actually aims at solving a robotics problem
but which relies on the semantic understanding of the environment. Last but not least,
source code from computer vision research related to semantic segmentation
is often made available, which is a great achievement. Each research group, however,
uses a different framework and adapting the trained networks to an own robotics 
codebase can sometimes take a considerable amount of development time.

\begin{figure}[t]
\vspace{2mm}
\centering
\setlength\tabcolsep{0.008\linewidth}
\begin{tabular}{cc}
  \includegraphics[width=0.47\linewidth]{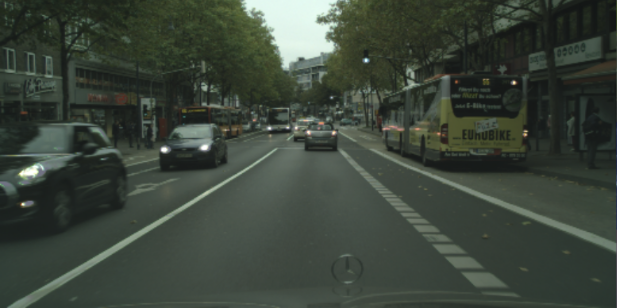} &
  \includegraphics[width=0.47\linewidth]{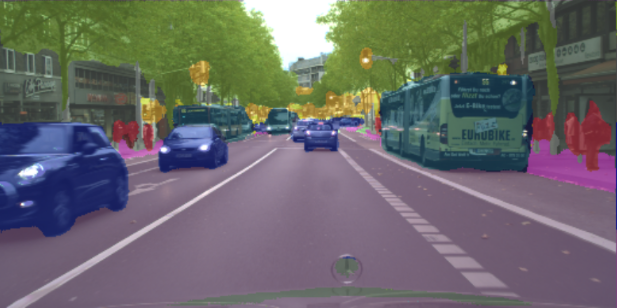} \\
  \includegraphics[width=0.47\linewidth]{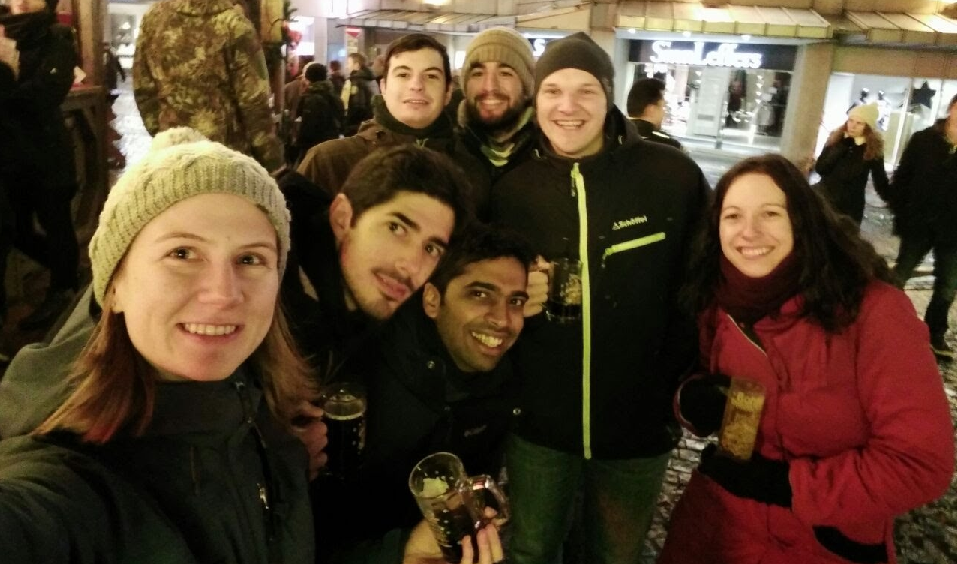} &
  \includegraphics[width=0.47\linewidth]{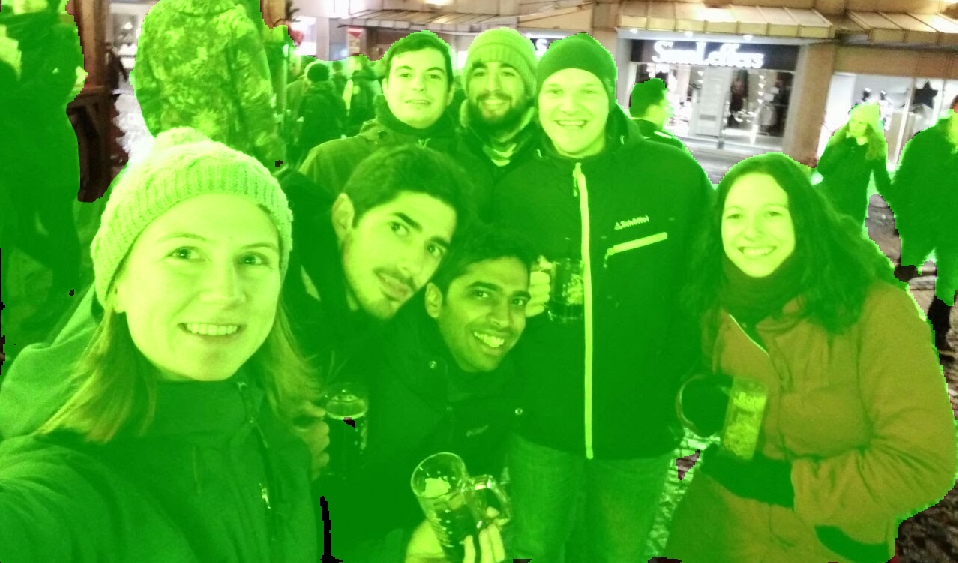} \\
  \includegraphics[width=0.47\linewidth]{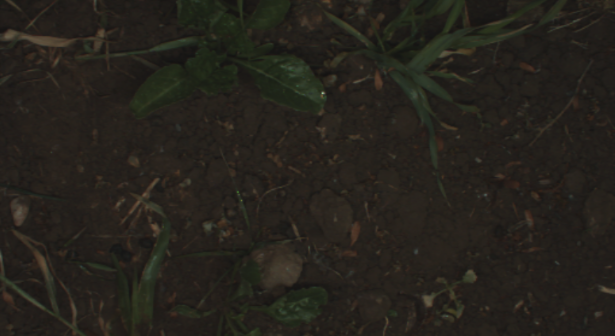} &
  \includegraphics[width=0.47\linewidth]{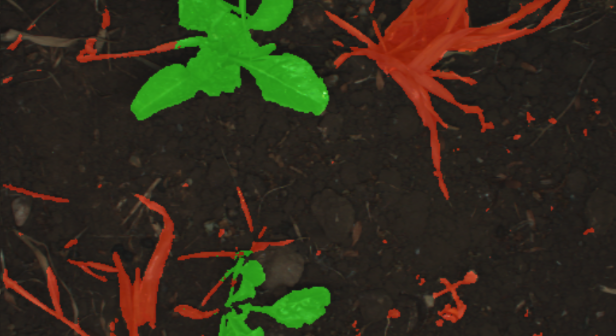}
\end{tabular}
\caption{Sample predictions from Bonnet. Left: Raw RGB images.
Right: Overlay with semantic segmentation label from CNN prediction.
From top to bottom: Cityscapes dataset \cite{cordts2016cvpr}, 
person segmentation inferring a photo from our research group,
trained on COCO~\cite{lin2014eccv}, Crop-Weed agricultural dataset \cite{chebrolu2017ijrr}.
Best viewed in color.}
\label{fig:samples_front_page}
\end{figure}

%% HOW & WHAT  
Therefore, we see the need for a tool that allows a developer to
easily train and deploy semantic segmentation networks for robotics.
Such a tool should  allow developers
to easily add new research approaches into the robotic system while avoiding the
effort of re-implementing them from scratch or modifying the available code
until it becomes at least marginally usable for the research purpose. This is
something that we experienced ourselves and observed in the community too often.

%% MAIN CONTRIBUTION & WHAT FOLLOWS FROM THAT

The contribution of this paper is a stable, easy to use, software tool with a modular
codebase which implements semantic segmentation using CNNs. It solves training and
deployment on a robot. Thus, we do not propose a new CNN approach here. Instead,
we provide a clean and extensible implementation to make this technology easily usable in
robotics and to enable a larger number of people to use CNNs for semantic
segmentation on their robots.
We strongly believe that our tool allows the scientific robotics
community to save time on the CNN implementations, enabling researchers to spend 
more time to  focus on how such information can aid robot perception, localization, mapping, path planning, obstacle avoidance, manipulation, safe navigation, etc. We show this with
different example use cases from the community, where robotics researchers with
no expertise in deep learning were able to, using Bonnet, train and deploy semantics in their
systems with minimal effort.
Bonnet relies on TensorFlow for our graph definition and training, but provides the possibility of using different backends with a clean and stable C++ API for deployment.
It allows for the possibility to transparently exploit custom hardware accelerators that
become commercially available, without modifying the robotics codebase.

%% CLAIMS (can be merged with the main contribution above)

In sum, we provide
(i) a modular implementation platform for training and deploying semantic segmentation CNNs in robots;
(ii) three sample architectures that perform well for a variety of perception problems 
in robotics while working roughly at sensor framerate;
(iii) a stable, easy to use, C++ API that also allows for the addition of new hardware 
accelerators as they become available;
(iv) a way to promptly exploit new datasets and network architectures as they are introduced
by computer vision and robotics researchers. 

Although we do not propose a new scientific method, we believe that this work has
a strong positive impact on the robotics community.
Six months after becoming publicly available, Bonnet already has a considerable
user base and won ``Best Demo Award" at the Workshop on Multimodal Robot Perception 
at ICRA 2018.
Our open-source software is available at \url{https://github.com/PRBonn/bonnet}.

%%%%%%%%%%%%%%%%%%%%%%%%%%%%%%%%%%%%%%%%%%%%%%%%%%%%%%%%%%%%%%%%%%%%%%%%%%%%%%%%
\section{Related Work}
\label{sec:related}

Semantic segmentation is important in robotics. The pixel-wise prediction of
labels can be precisely mapped to objects in the environment and thus allowing
the autonomous system to build a high resolution semantic map of its
surroundings.

One of the pioneers in efficient feed-forward encoder-decoder approaches to semantic segmentation 
is Segnet~\cite{badrinarayanan2015pami}. It  uses an encoder based on VGG16~\cite{simonyan2014arxiv},
and a symmetric decoder outputting a semantic label for each pixel of the input image. The decoder
uses the encoder pooling indexes to perform the unpooling to recover
some of the lost spatial resolution during pooling. Segnet is available as a Caffe
implementation and has pre-trained weights for several datasets. 
U-Net~\cite{ronneberger2015arxiv}, which was released contemporaneously, exploits the same 
encoder-decoder architecture but uses a decoder concatenation of the whole encoder 
feature map instead of sharing pooling indexes. This allows for  more accurate 
decision boundaries, which comes at a higher computational and memory cost. U-Net is available as
an implementation in a modified Caffe version and provides pre-trained weights for a medical dataset.
PSP-Net~\cite{zhao2016arxiv} uses ResNet~\cite{he2016cvpr} as the encoder, and exploits global information through a pyramid of average-pooling layers
after the latter, to provide more accurate semantics based on the environment of the image
objects. PSP-Net is also available as a modified Caffe implementation and comes with pre-trained weights from different
scene parsing datasets. All of these architectures are based on encoders such as
VGG and ResNet, which focus on accuracy of the predictions  rather than the execution speed 
for a near real-time  application in robotics.

Other architectures use post-processing steps  to improve the 
decision boundaries in the segmented masks. Some versions of DeepLab~\cite{chen2016arxiv} use
fully connected conditional random fields~(CRF) in addition to the last layer
CNN features in order to improve the localization performance. CRF-as-RNN~\cite{zheng2015iccv}
replaces the CRF with a recurrent neural network for prediction refinement, also
deviating from a fully feed-forward implementation. Both approaches provide modified 
implementations of Caffe and pre-trained weights for some scene parsing datasets.
Because of  rather inefficient feature extractors and the post-processing steps,
their execution speed is quite far away from the frame-rate of a regular camera, even when executed on the most powerful acceleration hardware available today.

Robots, however, need online inference capabilities for most applications. There has been work focusing
on inference efficiency, both in terms of execution time and model size. Enet~\cite{paszke2016arxiv}
proposes efficient down-sampling modules, efficient bottlenecks, and dilated convolutions
to decrease the model size and to improve the computational efficiency. Enet is available as an Torch implementation and provides pre-trained weights.
ICNet~\cite{zhao2017arxiv} proposes a compressed pyramid scene parsing network using an
image cascade that incorporates multi-resolution branches  to
provide a more efficient implementation of PSP-Net that can run closer to real-time. It is available
as a Caffe implementation based in PSP-Net, and contains pre-trained weights.
ERFNet~\cite{romera2018tits} proposes a way of widening each layer by replacing
the bottleneck modules with efficient dilated separable convolution modules. It is available 
both, as Torch and PyTorch implementations, and contains pre-trained weights. 
Mobilenets-v2~\cite{Sandler2018arXiv} proposes inverted residuals and linear bottlenecks to achieve near state-of-the-art performance in semantic segmentation using efficient constrained networks. Mobilenets-v2 is available as a TensorFlow implementation.

This fragmentation of different systems and backends motivates our idea of providing 
a modular implementation tool, in which such architectures can be realized. 

\section{Bonnet: Training and Deployment \\for Semantic Segmentation in Robotics}

\begin{figure}[t]
\vspace{2mm}
  \centering
  \begin{tabular}{c}
    {\includegraphics[width=0.95\linewidth]{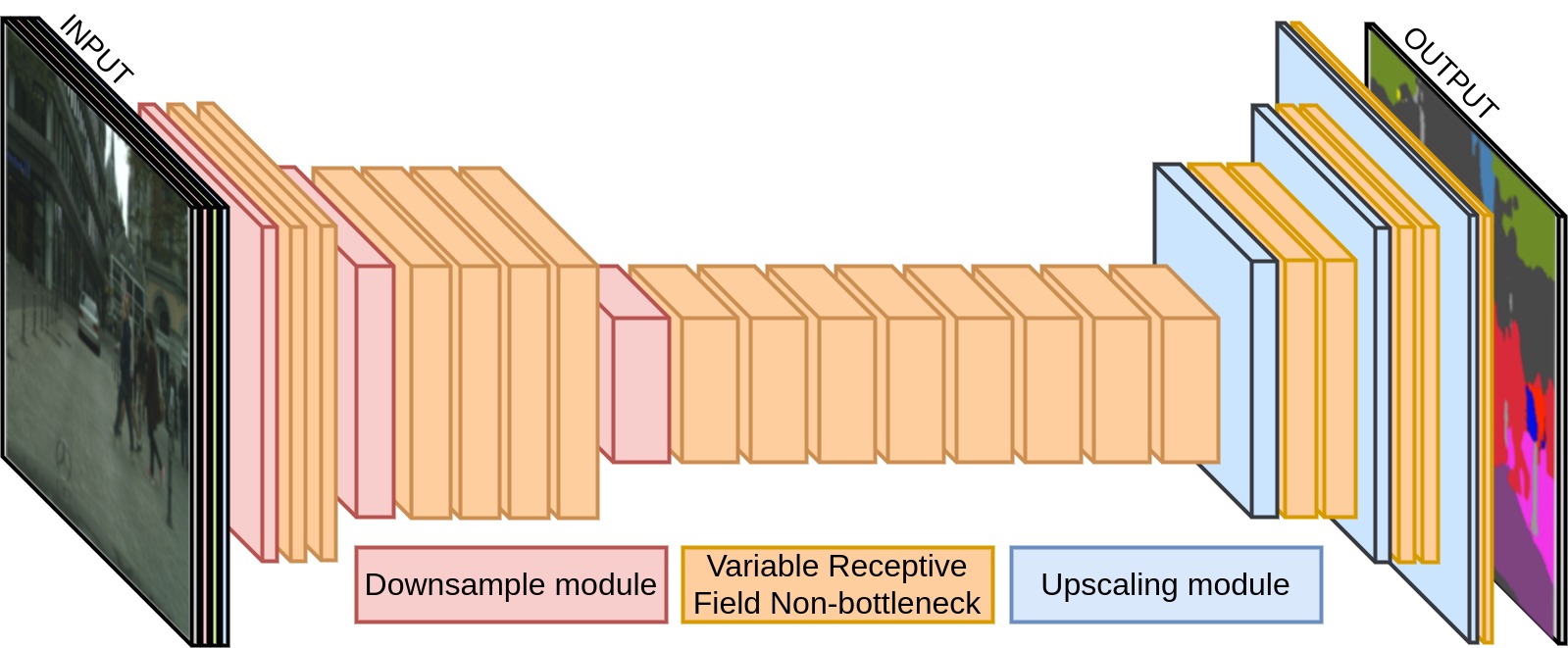}} \\
  \end{tabular}
  \caption{Example of an encoder-decoder semantic segmentation CNN implemented in
  Bonnet. It is based on the non-bottleneck idea behind ERFNet~\cite{romera2018tits}.
  Best viewed in color.}
  \label{fig:3dcnn}
\end{figure}

We provide our semantic segmentation tool called Bonnet with a Python training pipeline and a 
C++ deployment library. The C++ deployment library can be used standalone or as a ROS node. 
We provide three sample architectures focusing on realtime inference, based of
ERFNet~\cite{romera2018tits}~(see~\figref{fig:3dcnn}), InceptionV3~\cite{Szegedy2015arxiv}, and MobilenetsV2~\cite{Sandler2018arXiv} as well as pre-trained weights on four different datasets.
Our codebase allows for fast multi-GPU
training, for easy addition of new state-of-the-art architectures and available
datasets, for easy training, retraining, and deployment in a robotic system. It furthermore
allows for transparently using different backends for hardware accelerators as they
become available. This all comes with a stable C++ API.

\begin{figure}[t]
\vspace{2mm}
  \centering
  \begin{tabular}{c}
    {\includegraphics[width=0.5\linewidth]{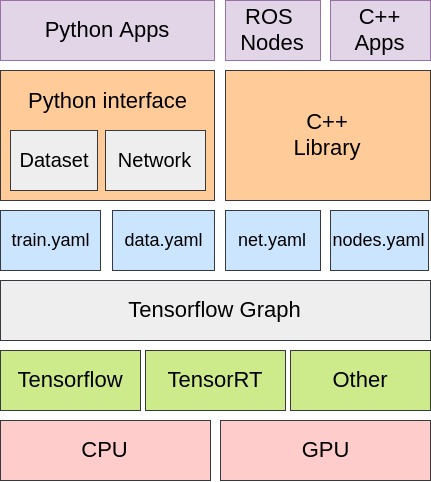}} \\
  \end{tabular}
  \caption{Abstraction of the codebase. Python interface is used for training and
  graph definition, and C++ library can use a trained graph and infer semantic
  segmentation in any running application, either linking it or by using the ROS
  node. Both interfaces communicate through the four configuration
  files in  \textit{yaml} format and the  trained model weights.}
  \label{fig:tool}
\end{figure}

The usage of Bonnet is split in two steps.
First, training the models to infer the pixel-accurate semantic classes from a specific dataset
through a Python interface which is able to access the full-fledged API provided by
TensorFlow for neural network training. Second,
deploying the model in an actual robotic platform through a C++ interface which 
allows the user to infer from the trained model in either an existing C++ application or 
a ROS-enabled robot.
\figref{fig:tool} shows a modular description of this division, from the 
application level to the hardware level, which we explain 
in detail in the following sections.
Note that for a reasonable number of use-cases, a developer using Bonnet can avoid
coding more or less completely. By simply providing own training data, a new application
can be deployed in a robot by simply fine-tuning one of the models and deploying using the 
ROS node.

\section{Bonnet Training}

The training of the models is performed through the methods defined through the 
abstract classes \class{Dataset} and \class{Network} (see~\figref{fig:tool}), which handle the pre-fetching, randomization,
and pre-processing of the images and labels, and the supervised training of the CNNs,
respectively. 

In order to train a model using our tool, there is a sequence of well-defined
steps that need to be performed, which are:

\begin{itemize}
  \item Dataset definition, which is optional if the dataset is provided in one 
        of our defined standard dataset formats.
  \item Network definition, which is also optional if the provided architecture
        fits the needs of the addressed semantic segmentation task.
  \item Hyper-parameter tuning.
  \item GPU training, either through single or multiple GPUs. This step can be performed
        either from scratch, or from a provided pre-trained model.
  \item Graph freezing for deployment, which optimizes the models to strip them
        from training operations and outputs a different optimized model format 
        for each supported hardware family.
\end{itemize}

\subsection{Dataset Definition}
The abstract class \class{Dataset} provides a standard way to access dataset files,
given a desired split for it in training, validation, and testing sets. The codebase contains
a general dataset parser, which can be used to import a directory containing images and
labels that are split into our standard dataset format. This parser can also be used as
a guideline to implement an own parser, for an own organization of the dataset files.
The definition of each semantic class, the colors for the debugging masks, the desired image inference size, and the location of the dataset are meant to be performed in the corresponding dataset's \textit{data.yaml} configuration file, 
of which there are several examples in the codebase. 
Once the dataset is parsed into the standard format, the abstract class \class{Network}
 knows how to communicate with it in order to handle the training and inference of the model.
Besides the handling of the file opening and feeding to the CNN trainer, the abstract 
dataset handler performs the desired dataset augmentation, such as flips, rotations, shears, stretches,
and gamma modifications. The dataset handler runs on a thread different from the training, such
that there is always an augmented batch available in RAM for the network to use, but also
allows the program to use big datasets in workstations with limited memory. The selection of this
cache size allows for speed vs. memory adjustment, which depend on the system available to the trainer.

\subsection{Network Definition}
Once the dataset is properly parsed into the standard format, the CNN
architecture has to be defined. We provide three sample architectures and provide
pre-trained weights for different datasets, and different network sizes, 
depending on the complexity of the problem. 
Other network architectures can be easily added, given the 
modular structure of our codebase, and it is the main purpose of the tool to
allow the implementation of new architectures as they become available. For this,
the user can simply create a new architecture file, which inherits the abstract \class{Network} class,
and define the graph using our library of layers. If a novel layer needs to be added,
it can be implemented using TensorFlow operations.
The abstract class \class{Network}, see \figref{fig:tool}, contains the definition of the training method that handles
the optimization through stochastic gradient descent, inference methods to test
the results, metrics for performance assessment, and the graph definition method, which
each architecture overloads in order to define different models. If a new architecture requires
a new metric or a different optimizer, these can be modified simply by overloading the corresponding method of the abstract class. 
The interface with the model architecture is done through the 
\textit{net.yaml} configuration file, which includes the selection of the architecture, the number of layers, number of kernels per layer, and some other architecture dependent hyper-parameters such as the amount of dropout~\cite{hinton2012arxiv}, and the 
batch normalization~\cite{ioffe2015arxiv} decay. 

The interface with the optimization is
done through the \textit{train.yaml} configuration file, which contains all training hyper-parameters, such as
learn rate, learn rate decay, batch size, the number of GPUs to use, and some other parameters
such as the possibility to periodically save image predictions for debugging, and summaries of 
the weights and activations histograms, which take a lot of disk space during training, and are only
useful to have during hyper-parameter selection. There are examples of these configuration files
provided for the included architectures in the codebase.

It is important to notice that since the abstract classes \class{Network} and \class{Dataset} handle
most cases well with their default implementation, no coding is required to 
add a new task and train a model unless for special cases. However, if a
complex dataset is to be added, or a new network implementation is desired, Bonnet allows
for its easy implementation.

\subsection{Hyper-parameter Selection}
Once the network and the dataset have been properly defined, the hyper-parameters need
to be tuned. We recommend doing the hyper-parameter selection through random-search,
as single GPU jobs, which can be performed by starting the training with different configuration files
(\textit{net.yaml}, \textit{train.yaml}), with all summary options enabled, and then choosing
the best performing model for a final multi-GPU training until convergence. The tool is designed
in this way for more simplicity, and because the hyper-parameter selection 
jobs can be scheduled easily with an external job-scheduling tool. 
Some of the hyper-parameters which can be configured are: the number of images to cache in RAM,
the amount and type of data augmentation, the decays for batch normalization~\cite{ioffe2015arxiv}
and regularization through weight decay and dropout~\cite{hinton2012arxiv}, the learning rate and
momentums for the optimizer, the type of weighting policy for dealing with unbalanced classes in the dataset, the $\gamma$ for the focal loss~\cite{lin2017arxiv}, the batch size, and number of GPUs.

\begin{figure}[t]
\vspace{2mm}
  \centering
  \begin{tabular}{c}
    {\includegraphics[width=0.55\linewidth]{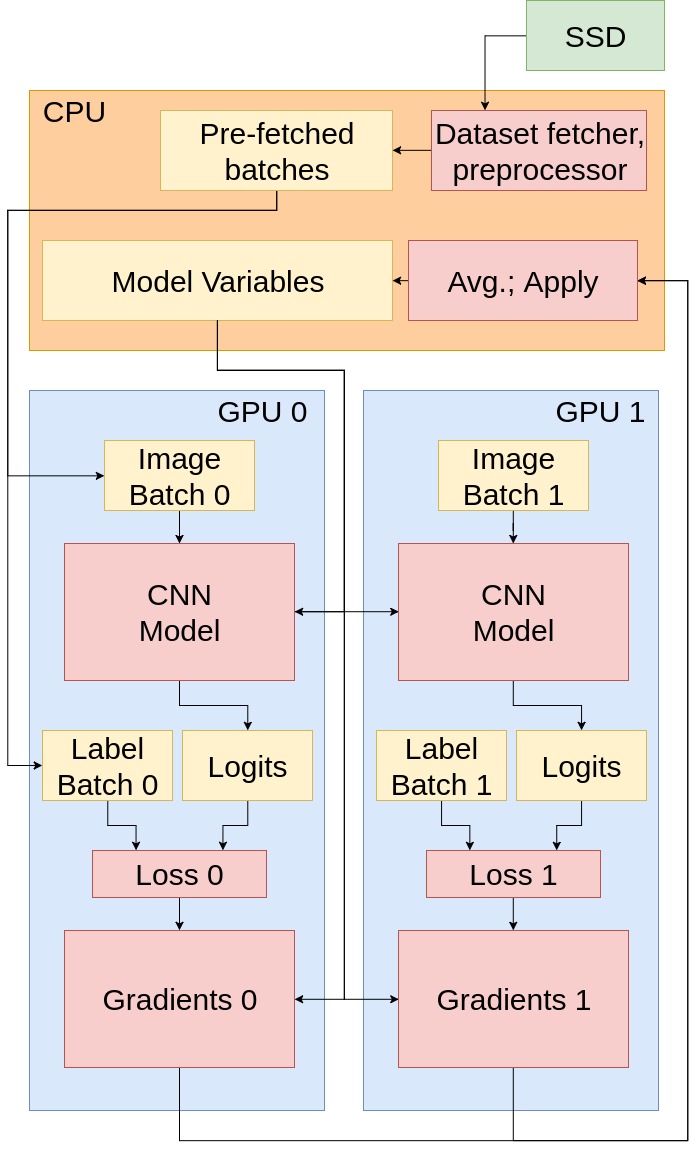}} \\
  \end{tabular}
  \caption{Multi-GPU training. Example using two GPUs, but scalable to all GPUs
  available in workstation.}
  \label{fig:multigpu}
\end{figure}

\subsection{Multi GPU training}

Once the most promising model is found, the training can be done with this hyper-parameter set using
multiple GPUs to be able to increase the batch size, and hence, the speed of training. 
Changing the number of GPUs used for training is as simple as changing the setting in
the \textit{train.yaml} configuration file, but we recommend scaling the hyper-parameter set
found following the procedure described in~\cite{goyal2017arxiv} for better results. 
The multi-GPU training, as described in \figref{fig:multigpu}, is performed by synchronously
averaging the gradients obtained by a single Stochastic Gradient Descent step in each GPU. For this, all model parameters
are stored in main memory and they are transferred to each GPU after each step of averaged gradient
update. This is handled by the abstract network's training method, and it is transparent to the user.
The accuracy and Jaccard index (IoU) are periodically reported and the best performing models 
in the validation set are stored. We store both the best accuracy and the best intersection over union
model, for posterior use in deployment. The mean Jaccard index (IoU) is used for the final evaluation:
$${mIoU = \frac{1}{C}~\sum_{i=1}^{C}\frac{\text{TruePos}_j}{\text{TruePos}_j + \text{FalsePos}_j + \text{FalseNeg}_j}}$$

Another important work to make GPU training more efficient is the introduction
of the concept of ``checkpointed gradients"~\cite{chen2016arxivb}, which allows to
fit big models in GPU memory in sub-linear space. This is done by checkpointing nodes
in the computation graph defined by the model, and recomputing the parts of the 
graph in between those nodes during backpropagation. This makes it possible to calculate
the network gradients in the backward pass at reduced memory cost, without increasing the 
computational complexity linearly. Our tool allows to use the implementation of
the checkpointed gradients, and therefore, besides allowing for bigger batches due to the
multi-GPU support, it also allows for bigger per-GPU batches.

\subsection{Graph Freezing for Deployment}
Once the trained model performs as desired, the tool exports a log directory
containing a copy of all the configuration files used, for later reference, and 
two directories inside containing the best IoU and best accuracy checkpoints. To
deploy the model and use it with different back-ends, such as TensorRT, we need
to ``freeze" the desired model. Freezing removes all of the helper operations 
required for training and unnecessary for inference, such as the optimizer ops, the gradients,
dropout, and calculation of train-time batch normalization momentums.
The abstract network provides a method which handles this procedure and
creates another directory with four frozen models: the model in NCHW format, which is faster
when inferring using GPUs; the model in NHWC format, which can be faster when using CPUs;
an optimized model, which tries to further combine redundant operations, and an 8-bit
quantized model for faster inference. This method also generates a new configuration file called
\textit{nodes.yaml}, which contains important node names, such as the inputs, code, and
outputs as logits, softmax, and argmax. This allows for a more automated parsing of the frozen
model during inference and automatically remembering the names of the inputs and outputs. We provide a Python script
for this procedure, which takes a training log directory as an input and outputs all the frozen models
and their configuration files in a packaged directory that contains all files needed for deployment.
We also provide other applications to test this model in images and videos, in order to observe
the performance qualitatively for debugging, and to serve as an example for serving using python, in
case this is desired. It is key to notice that since the whole process can be performed
in a host PC, the device PC on the robot only needs the dependencies to run the inference,
such as our C++ library.

\begin{figure}[]
\begin{minipage}{\linewidth}
\lstset{language=[GNU]C++,
        basicstyle=\small,
        keywordstyle=\color{blue}\ttfamily,
        stringstyle=\color{red}\ttfamily,
        commentstyle=\color{gray}\ttfamily,
        morecomment=[l][\color{magenta}]{\#},
        escapeinside={<@}{@>}
}
\begin{lstlisting}[caption={\label{lst:snippet}C++ code showing simplicity of semantic segmentation CNN
inference in C++ application, using Bonnet tool as a library.}]

#include <bonnet.hpp>
#include <opencv2/core/core.hpp>
#include <string>

int main() {
  // path to frozen dir
  std::string path = "/path/frozen/pb";
  // tf for Tensorflow, trt for TensorRT
  std::string backend = "trt";
  // gpu or cpu (or specialized)
  std::string dev = "/gpu:0";

  // Create the network
  bonnet::Bonnet net(path, backend, dev);
  // Infer image from disk
  cv::Mat image, mask, mask_color;
  image = cv::imread("/path/to/image");
  net.infer(image, mask);
  // If necessary, colorize (like Fig.1)
  net.color(mask, mask_color);
}
\end{lstlisting}
\end{minipage}
\end{figure}

\begin{table}[t]
\vspace{4mm}
\centering
\caption{Pixel-wise metrics for sample architectures.}
{
\footnotesize
\setlength\tabcolsep{3pt}
\begin{tabular}{ccccccc}
\hline
Dataset & Arch. & Input Size & \#Param &\#Ops. & mIoU & mAcc. \\
\hline
\multirow{3}{*}{Cityscapes} & ERFNet & 1024x512 & 1.8M & 66B & 62.8\% &  92.7\%\\
 & Mobilenets & 768x384 & 6.9M & 72B & 63.5\% &  93.7\%\\
 & Inception & 768x384 & 4.3M & 47B & 66.4\% &  94.1\%\\
\hline
COCO & \multirow{2}{*}{Inception} & 640x480 & \multirow{2}{*}{4.3M} & 48B & 87.1\% & 97.8\%\\
Persons & &  320x240 &  & 12B & 83.4\% &  96.9\%\\
\hline
\multirow{2}{*}{Synthia} & \multirow{2}{*}{ERFNet} & 512x384 & \multirow{2}{*}{1.8M} & 24B & 64.1\% & 92.3\% \\
& &  960x720 &  & 85B & 71.3\% &  95.2\%\\
\hline
Crop-Weed & ERFNet & 512x384 & 1.1M & 9B & 80.1\% & 98.5\% \\
\hline
\end{tabular}
}
\label{tab:acc}
\end{table}

\section{Bonnet Deployment}
For the deployment of the model on a real robot, we provide a C++ library with an abstract 
handler class that takes care of the inference of semantic segmentation, and allows for each
implemented back end to run without changes in the API level.
The library can handle inference from a frozen model that is generated
through the last step of the Python interface. Bonnet handles the inference
through the user's selection of the desired back end, execution device~(GPU, CPU, or other
accelerators), and the frozen model to use. There are two ways to access this
library. One is by linking it with an existing C++ application, using the two 
provided standalone applications as a usage example. The second one is to use the provided
ROS node, which already takes care of everything needed to do the inference, from
debayering the input images, to resizing, and publishing the mask topics, so that no
coding is needed. \listref{lst:snippet} contains an example of how to build a small
``main.cpp" application to perform semantic segmentation on an image from disk
using our C++ library.

\section{Sample Use Cases Shipped with Bonnet}

In order to show the capabilities of Bonnet, we provide three sample architectures
focusing on realtime inference. The three models included are based on
ERFNet~\cite{romera2018tits}, InceptionV3~\cite{Szegedy2015arxiv}, and MobilenetsV2~\cite{Sandler2018arXiv}, with minor modifications which allow to run the architectures
in TensorRT, which supports a subset of all TensorFlow operations, and makes the
networks much faster to run, as we show in \tabref{tab:speed}.

\tabref{tab:acc} shows the performance of the sample architectures on four diverse and challenging
datasets, two for scene parsing, one for people segmentation, and one for precision agriculture purposes, 
for which we provide the trained weights. 
Because each problem presents a different level of difficulty and uses images of a different
aspect ratio, we show the performance of the model for different number of parameters and number
of operations by varying the number of kernels of each layer of the base architecture and the size of the input.

\begin{table}
\vspace{4mm}
\centering
\caption{Mean runtime of the erfnet-based architecture for different datasets, input
sizes, and backends.}
{
\footnotesize
\setlength\tabcolsep{3pt}
\begin{tabular}{ccccc}
\hline
Dataset & Input Size & Back-end & GTX1080Ti & Jetson TX2 \\
\hline
\multirow{4}{*}{Cityscapes} & \multirow{2}{*}{512x256} & TensorFlow & 19ms~(52\,FPS) & 170ms~(6\,FPS) \\
& & TensorRT & 10ms~(100\,FPS) & 89ms~(11\,FPS) \\
& \multirow{2}{*}{1024x512} & TensorFlow & 71ms~(14\,FPS) & 585ms~(2\,FPS) \\
& & TensorRT & 33ms~(30\,FPS) & 245ms~(4\,FPS) \\
\hline
& \multirow{2}{*}{640x480} & TensorFlow & 27ms~(37\,FPS)  & 321ms~(3\,FPS) \\
COCO & & TensorRT & 15ms~(65\,FPS) & 128ms~(8\,FPS) \\
Persons& \multirow{2}{*}{320x240} & TensorFlow & 21ms~(47\,FPS) & 200ms~(5\,FPS) \\
&  & TensorRT & 7ms~(142\,FPS) & 80ms~(14\,FPS) \\
\hline
\multirow{4}{*}{Synthia}  & \multirow{2}{*}{512x384} & TensorFlow & 20ms~(50\,FPS)  & 223ms~(4\,FPS) \\
 & & TensorRT & 11ms~(100\,FPS) & 127ms~(8\,FPS) \\
 & \multirow{2}{*}{960x720} & TensorFlow & 61ms~(16\,FPS) & 673ms~(1\,FPS) \\
  &  & TensorRT & 27ms~(37\,FPS) & 362ms~(3\,FPS) \\
\hline
\multirow{2}{*}{Crop-Weed}  & \multirow{2}{*}{512x384} & TensorFlow & 9ms~(111\,FPS)  & 132ms~(8\,FPS) \\
 & & TensorRT & 4ms~(250\,FPS) & 99ms~(10\,FPS) \\
\hline
\end{tabular}
}
\label{tab:speed}
\end{table}
Since Bonnet is meant to serve as a general starting point to implement different architectures,
we advise referring to the code in order to have an up-to-date measure of the latest architecture
design performances.

\tabref{tab:speed} shows the runtime of the ERFNet based model, with varying complexity 
and input size. It shows how much the inference time can be improved by using custom
accelerators for the available commercial hardware. This further supports the importance of allowing
the user to transparently benefit from its usage with no extra coding effort, as well as providing
a modular C++ backend which allows the support of other backends as they become
available.

\section{Sample Use Cases From The Community}

\begin{figure}[t]
\vspace{2mm}
\centering
\setlength\tabcolsep{0.008\linewidth}
\begin{tabular}{cc}
  \multicolumn{2}{c}{\footnotesize{(a)~Dynamic object removal for mapping}} \\
  \includegraphics[width=0.44\linewidth]{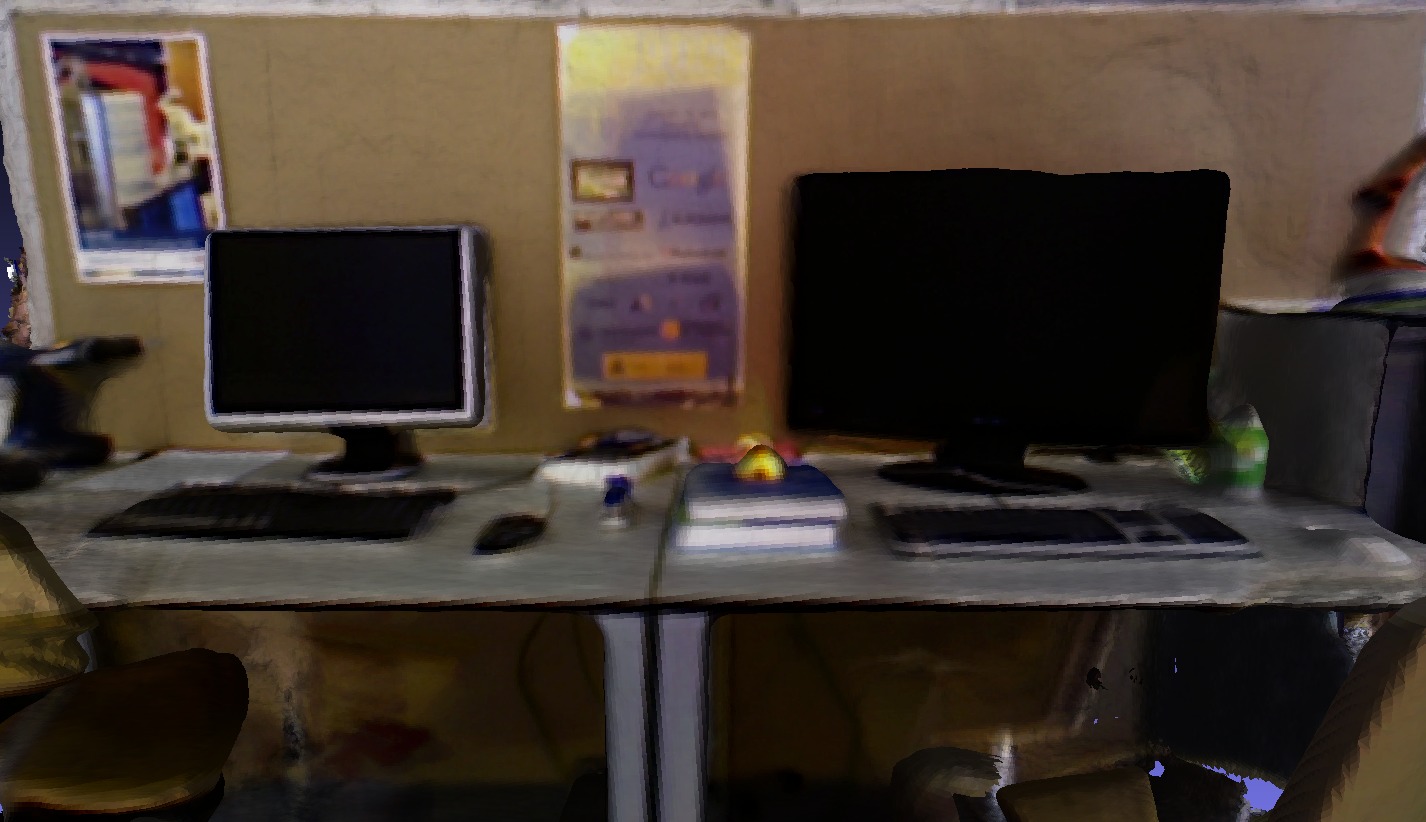} & 
  \includegraphics[width=0.44\linewidth]{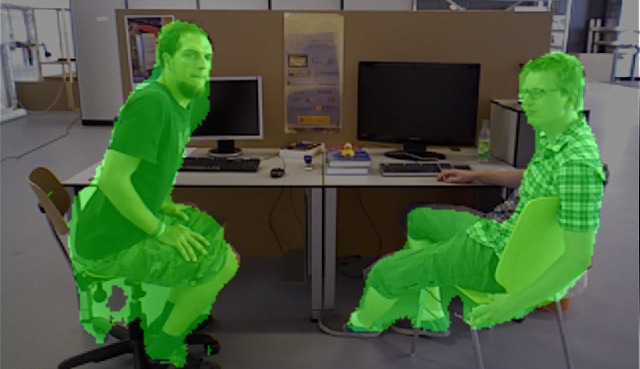} \\
  \multicolumn{2}{c}{\footnotesize{(b)~Berry detection in wineyards}} \\[1mm]
  \includegraphics[width=0.44\linewidth]{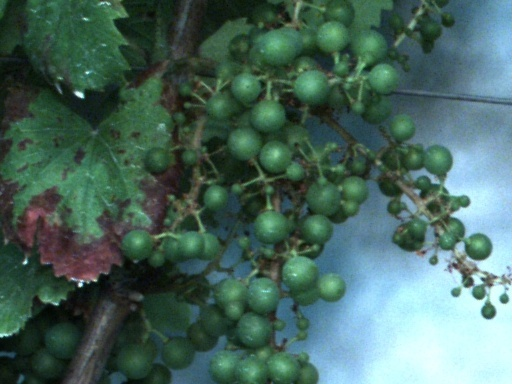} &
  \includegraphics[width=0.44\linewidth]{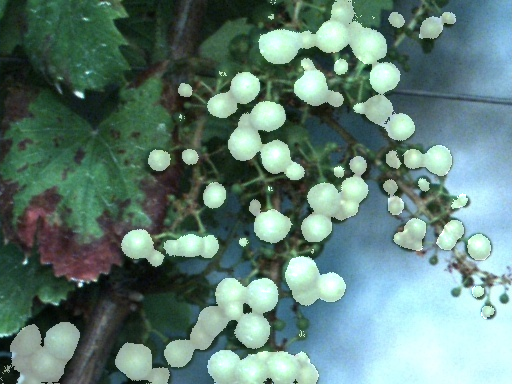} \\
  \multicolumn{2}{c}{\footnotesize{(c)~Facade segmentation}} \\[1mm]
  \includegraphics[width=0.44\linewidth]{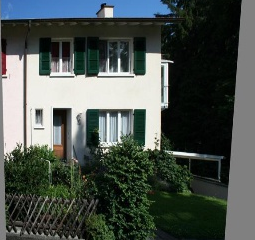} &
  \includegraphics[width=0.44\linewidth]{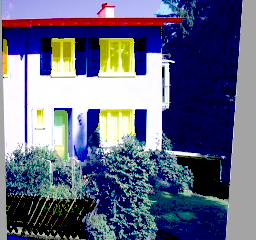} \\
  \multicolumn{2}{c}{\footnotesize{(d)~Toy segmentation for efficient path planning of a humanoid}} \\[1mm]
  \includegraphics[width=0.44\linewidth]{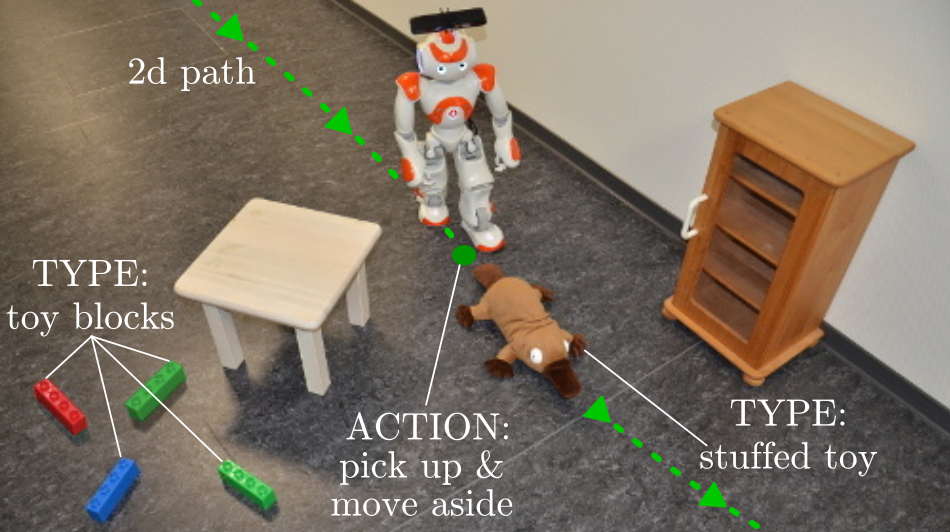} &
  \includegraphics[width=0.44\linewidth]{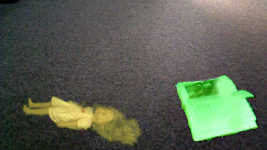} 
\end{tabular}
\caption{Sample use cases from the community. Left: illustration of the application.
Right: overlay RGB image with semantic prediction from Bonnet. In (a), the left column
represents the generated 3D model filtering the dynamics, not the input image.}
\label{fig:samples_community}
\end{figure}

\figref{fig:samples_community} section shows some example use cases from
other robotics researchers where one of the architectures was used with the
standard parser to train and deploy Bonnet semantics in four different applications,
with zero coding effort, from training to deployment using C++ or ROS.
Use case (a) uses our person segmentation trained on COCO
and the C++ library as an off-the-shelf preprocessing tool to remove dynamics from
camera data before feeding it into a TSDF-based GPU-accelerated realtime mapping
pipeline. In (b), an inception-based model was trained to recognize berries in
wine yards for automated, robotic, yield estimation. In (c), the ERFNet model
was retrained starting from Cityscapes weight in order to infer the segmentation
of facade elements using the ETRIMS dataset~\cite{forstner2009etrims}. Finally,
in (d), the inception-based model was trained to recognize toys using a large
database of objects downloaded from the Internet, and deployed using the ROS node
in a humanoid robot with a JetsonTX2 for efficient, semantic, path planning~\cite{regier2018humanoids}. Bonnet
has been used in several other use cases by the community.

%%%%%%%%%%%%%%%%%%%%%%%%%%%%%%%%%%%%%%%%%%%%%%%%%%%%%%%%%%%%%%%%%%%%%%%%%%%%%%%%
\section{Conclusion}
\label{sec:conclusion}
In this paper, we presented Bonnet, an open-source semantic segmentation training
and deployment tool for robotics research. Bonnet eases the
integration of semantic segmentation methods for robotics. It provides
a stable interface allowing the community to better collaborate,  add different datasets 
and network architectures, and share  implementation efforts as well as pre-trained models.
We believe that this tool speeds up the deployment of semantic segmentation CNNs 
on research robotics platforms.
We provide three sample architectures that operate at framerate, and include
pre-trained weights for diverse and challenging datasets with the goal that the
robotics community will exploit them and contribute to the tool.

%%%%%%%%%%%%%%%%%%%%%%%%%%%%%%%%%%%%%%%%%%%%%%%%%%%%%%%%%%%%%%%%%%%%%%%%%%%%%%%%
% Future work only if it makes sense 
%\emph{Future work: Use only if applicable -- but if so, use the following sentence to start:} 
%
%Despite these encouraging results, there is further space for
%improvements. For example, ...

%%%%%%%%%%%%%%%%%%%%%%%%%%%%%%%%%%%%%%%%%%%%%%%%%%%%%%%%%%%%%%%%%%%%%%%%%%%%%%%%
% Only if applicable
\section*{Acknowledgments}
We thank Laura Zabawa, Susanne Wenzel, Emanuele Palazzolo, and Peter Regier for
for useful feedback and for providing the images with example use cases of Bonnet for their current research.

\clearpage
\bibliographystyle{plain}

% All new citations should go to new.bib. The file glorified.bib should go
% be the one from the ipb server. After paper or related work has been 
% written merge the entries from new.bib to glorified.bib ON THE SERVER,
% replace the glorified.bib in this repository and empty the new.bib
\bibliography{glorified,new}

\end{document}